\documentclass[11pt]{article}
\usepackage{arxiv}   
\usepackage{graphicx}
\usepackage{booktabs}
\usepackage{amsmath}
\usepackage[numbers]{natbib}
\usepackage{hyperref}

\title{Train, Retrieve, or Both? A Four-Arm Head-to-Head for Correct Statutory
Citation on the Ontario Residential Tenancies Act}
\author{%
  Ali Asaria \\ Transformer Lab \and
  Tony Salomone \\ Transformer Lab \and
  Deep Gandhi\thanks{Corresponding author: \texttt{deep@lab.cloud}} \\ Transformer Lab
}
\date{}
\runningtitle{Train, Retrieve, or Both? Statutory Citation for the Ontario RTA}

\begin{document}
\maketitle

\begin{abstract}
Self-represented tenants, landlords, and help-desk staff need to be pointed at the
provision of law that actually governs a question, with a correct statutory citation.
We study this task on the Ontario \emph{Residential Tenancies Act, 2006} (RTA) and its
core regulation, asking the operator's question empirically: is fine-tuning enough, or is
hybrid retrieval needed? We run a four-arm head-to-head on Qwen2.5-7B-Instruct (base
zero-shot, LoRA SFT-only, RAG-only, and an SFT+RAG hybrid), scored on citation
exact-match (section+subsection) over a small, human-verification-pending real eval set.
The base model cannot cite the RTA and SFT-only mis-recalls sections; retrieval is essential and
drives hallucination to zero \emph{by construction}; and the \textbf{SFT+RAG hybrid scores
highest at 0.481 exact-match with zero hallucinated citations}. Its edge comes from SFT making
provision selection more robust to the higher-recall candidate sets that \emph{hurt} zero-shot
RAG. Notably, this cheap \texttt{bge-small} hybrid matches or beats a pipeline built on bigger,
specialized retrieval models (a larger embedder and a cross-encoder reranker), and a
larger/improved training set does not help either: strong statutory-citation performance here
does not require specialized retrieval models or more data. The artifact zeroes hallucination
and clears the lift-over-base bar but does \emph{not} reach the aspirational 0.70 exact-match
target. All results are on a small, human-verification-pending real eval set and are reported as
preliminary.
\end{abstract}

\section{Introduction}
Ontario's residential-tenancy law is governed by the \emph{Residential Tenancies Act, 2006}
(RTA, S.O.\ 2006, c.\ 17) and O.\ Reg.\ 516/06 (\emph{General}). The people who most need it
(self-represented tenants and landlords, paralegals, and tenant-help-desk staff) rarely
need fluent prose; they need to be pointed at \emph{the provision that governs}, with a
correct citation (the right Act/regulation, section, and subsection). We therefore treat the
headline quality target as the \textbf{citation score} rather than answer fluency: the unit
of prediction is a single natural-language tenancy question (e.g.\ ``How much notice must a
landlord give to enter my unit?'') mapped to a short answer plus one or more statutory
citations (e.g.\ \texttt{RTA s.\ 26(1)}). This is decision-support, not legal advice.

The operator posed a concrete, falsifiable question: \emph{is fine-tuning enough, or do we
need hybrid RAG?} Rather than assume an answer, we measure it. We build a citation index and
a question$\rightarrow$citation dataset (no public Ontario-RTA set exists) and run a
four-arm head-to-head on the same held-out citation metric: the base instruct model
zero-shot, LoRA supervised fine-tuning on synthetic question$\rightarrow$citation pairs,
RAG over a chunked statute index, and the SFT+RAG hybrid. Beyond a single-GPU LoRA budget, the stated design constraints
include interactive latency, a pinned consolidation date, and using no personal data
as model input.

\paragraph{Contributions.}
\begin{enumerate}
\item A measured four-arm answer to ``train vs.\ retrieve'': base cannot cite (0.00,
  81\% hallucinated), SFT-only mis-recalls sections (0.148), RAG is
  essential (0.44, zero hallucination), and the hybrid scores highest (0.481), giving the
  ordering base $\ll$ SFT-only $\ll$ retrieval-bearing arms; see \S\ref{sec:results}.
\item An \emph{SFT$\times k$ effect}: more retrieved candidates \emph{hurt} zero-shot RAG
  (0.44$\rightarrow$0.37 at $k{=}10$) but \emph{help} the SFT model
  (0.333$\rightarrow$0.481), so the hybrid gains selection robustness to higher-recall
  candidate sets; see \S\ref{sec:results}.
\item An efficiency result: the cheap \texttt{bge-small} hybrid matches or beats a pipeline
  built on bigger, specialized retrieval models (a larger embedder $+$ a cross-encoder
  reranker, 0.37), and a larger/improved training set does not improve over the original
  (0.444); see \S\ref{sec:results}.
\item An honest accounting: zero hallucinated citations \emph{by construction} and a
  $+0.48$ exact-match lift, but the 0.70 target is not met on a 27-item,
  human-verification-pending eval set; see \S\ref{sec:disc}.
\end{enumerate}

\section{Related Work}
\paragraph{Generating verifiable citations.}
Teaching attribution directly is more effective than passively tagging text. \citet{2506.17585v3}
show that bidirectional source$\leftrightarrow$fact training with fact-variation augmentation
lifts citation precision substantially, while verbatim token replay gives no benefit and more
replay epochs can hurt generalization, which directly motivates our paraphrase-based synthetic
generation and group-aware split. \citet{2501.17840v1} find continual pre-training on raw
documents has only a marginal effect, whereas reformatting to atomic facts yields large
exact-match gains, supporting SFT on \texttt{(question$\rightarrow$provision-id)} pairs over
dumping statute prose. \citet{2410.17952v2} self-generate QA from an unlabeled corpus with a
retrieval round-trip filter, and \citet{2603.13307v1} scale graph-derived synthetic data; both
inform our synthetic-from-statute recipe. Attribution methods such as TRACE
\citep{2407.04981v1} and judgement-citation retrieval by contextual similarity
\citep{2406.01609v2} frame citation as a retrieval/grounding problem rather than free
generation.

\paragraph{Retrieval for statute.}
Statute retrieval is semantics-dominant yet benefits from a lexical anchor:
\citet{2511.00268v1} report optimal fusion heavily weighted toward the dense retriever, but
\citet{2512.12117v1} and \citet{2508.00679v1} show BM25 is essential to anchor exact
identifiers and defined terms, with a robust fusion weight that transfers without per-corpus
tuning, so we adopt a hybrid BM25$+$dense default. Second-stage re-ranking
\citep{2511.00268v1} or cross-encoder rerankers \citep{2508.00679v1} can lift retrieval F1;
we test the reranker as an ablation. \citet{2504.16121v1} use relevance-check/query-refinement
loops and low decode temperature, and bounded-size chunking with overlap is
standard~\citep{2502.16573v1,2509.00141v1}; we instead chunk on statutory section
boundaries as a design choice, one provision per chunk. Crucially,
\citet{2603.13307v1} report that \emph{bad} retrieval actively hurts (over-retrieval misleads
generation), and \citet{2406.05365v2} note multi-stage pipelines compound failures; both
foreshadow our finding that heavier retrieval helps neither arm.

\paragraph{Anti-hallucination guardrails.}
\citet{2512.12117v1} use mechanical citation verification (accept a citation only if it exists
in the pinned corpus) to reach high accuracy with zero hallucinations; \citet{2512.01659v1}
reduce structural hallucination detection to a cheap entity-grounding existence check but warn
it collapses on short, sparse inputs, favoring a deterministic existence check for single
subsections. \citet{2406.05365v2} contrast large/small models to verify grounding, and
\citet{2509.26307v2} explore decode-time attribution; we adopt the cheap existence-check
guardrail as a hard gate. Reality checks abound: RAG does not eliminate hallucination
\citep{2508.00709v3}, commercial legal RAG tools still hallucinate 17--33\%
\citep{2512.01659v1}, and retrieval can game cheap metrics like ROUGE without improving judged
quality \citep{2508.00709v3}.

\paragraph{LoRA and positioning.}
Our central positioning is simple: \citet{2603.13307v1} ran three of our four arms and explicitly
left SFT+RAG as future ``Case D'', and this work fills that gap on a statutory-citation
exact-match metric, with a deterministic hallucination check that the surveyed legal papers
measured only qualitatively. On the training side, \citet{2512.15634v1} show LoRA usually
matches or beats full SFT while forgetting less, and that fine-tuning can erode a strong
base, arguing to keep the zero-shot base as a real baseline (which we do) and to prefer LoRA.

\section{Method}
\paragraph{Task and label.} For a question $q$ we predict a short answer and a set of citations
$\hat{C}=\{(\text{instrument}, \text{section}, \text{subsection})\}$ against a gold set $C$.
Correctness is section$+$subsection match; section-only matches earn partial credit (0.5); the
instrument must match. Citations to provisions absent from the pinned consolidation count as
hallucinations.

\paragraph{Corpus and index.} We parse the e-Laws consolidation of the RTA
(297 sections, 895 subsections, 20 Parts) and O.\ Reg.\ 516/06
(61 sections, 166 subsections) into a citation inventory of
358 provisions, each a single structure-aware RAG chunk. The inventory doubles as the
existence-check whitelist for the hallucination metric.

\paragraph{Arms.} The \textbf{base} arm is the instruct model zero-shot. The \textbf{SFT-only}
arm is a LoRA SFT on synthetic question$\rightarrow$citation pairs
(rank $r{=}32$, $\alpha{=}2r$, \texttt{lr}\,$2\mathrm{e}{-}4$, 400 steps;
$q/k/v/o$ projections), following the LoRA guidance of \citet{2512.15634v1}.
The \textbf{RAG-only} arm is the base model with hybrid BM25$+$\texttt{bge-small}
retrieval, prompted to cite only from the retrieved context (a cite-from-context guardrail that
is effectively an existence check). We note up front that this guardrail and the hallucination
\emph{metric} share the same pinned inventory, so a 0.00 hallucination rate for the
retrieval-bearing arms is largely \emph{guaranteed by construction} rather than learned;
we treat it as a design property, not an emergent result. The \textbf{SFT+RAG hybrid} is the SFT
model trained to select
the correct citation from retrieved candidates, evaluated with the same retrieval. We sweep
retrieval depth $k$ and retrieval strength (cheap \texttt{bge-small} vs.\ \texttt{bge-large}
$+$ cross-encoder reranker).

\paragraph{Objective.} The optimization target is the citation set, scored as exact-match with
section-only partial credit and multi-citation F1. Two reporting conventions matter. First,
because section-only matches earn 0.5, the ``exact-match'' figures we report are a
partial-credit citation score in $[0,1]$, not a strict 0/1 rate, so we keep the name
``exact-match'' for brevity but it is partial-credit-weighted. Second, F1 is computed per item
over the predicted vs.\ gold citation \emph{sets} (multi-citation precision and recall, then
their harmonic mean) and averaged across items; it can exceed exact-match because it gives
proportional credit for partially-correct citation sets. Conceptually, an arm's quality
combines retrieval supply and selection:
\begin{equation}
\text{ExactMatch} \;\le\; \underbrace{\mathrm{recall@}k}_{\text{retrieval supplies gold}}
\;\times\; \underbrace{P(\text{select gold}\mid\text{retrieved})}_{\text{SFT improves this}},
\end{equation}
which frames the central finding: SFT raises the selection term while RAG raises the supply
term, and the hybrid benefits from both.

\section{Experimental Setup}
\paragraph{Data.} No public Ontario-RTA question$\rightarrow$citation dataset exists, so s4
builds one. \textbf{Training (synthetic).} An LLM paraphrases each section/subsection into
natural questions with the provision as gold: 2{,}148 pairs, split leakage-safely (question-level
bigram-Jaccard dedup $+$ a 15\% unseen-section slice) into train 1{,}473 / synth\_test 345 /
unseen 330. \textbf{Held-out (real, headline).} A small set mined from
tenant-law sites with explicit RTA section references: 27 source-cited, in-inventory items,
each flagged human-verification-pending. Splits are group-aware by section/Part so the
model cannot win by memorizing question$\rightarrow$citation maps; the real set is fully
source-disjoint. Statute text is the e-Laws consolidation (Government of Ontario, Crown
copyright).

\paragraph{Protocol.} The base model is Qwen2.5-7B-Instruct (operator: Qwen-only;
Llama-3.1-8B was not evaluated, HF-gated). Metrics: citation exact-match
(instrument$+$section$+$subsection) with section-only partial credit, multi-citation
precision/recall/F1, a deterministic hallucinated-citation rate (existence check against the
pinned inventory), and recall@$k$ for the RAG arms to separate retrieval from generation
failure. The scorer is deterministic and reused across arms. A harness control (perfect
predictions on the 27 items $\rightarrow$ exact-match 1.0, hallucination 0.0) confirms the
scorer and gold are \emph{mechanically} correct: it validates the measurement instrument, not
the statistical reliability of the model numbers, which is bounded by $n{=}27$. All
arms are single-run point estimates (no seed replication). Total compute was
4.43 GPU-hours against a planned 44 (Lambda H100 for training, SkyPilot RTX 3090 for
eval).

\section{Results}
\label{sec:results}
Table~\ref{tab:results} reports the final model against the overview's success-criteria gates,
and Table~\ref{tab:arms} the full arm/ablation grid. All numbers are single-run point estimates
on the 27-item real eval (human-verification-pending) unless noted, so they should be read as
preliminary. On 27 items a single correct/incorrect prediction moves a rate by
$\sim$0.037, so we draw inferences only at the level of \emph{large}, qualitatively-robust gaps
(base vs.\ SFT-only vs.\ the retrieval-bearing arms); differences of one to three items among
the retrieval-bearing cells (e.g.\ 0.481 vs.\ 0.44, or the $k$ and retrieval-strength sweeps)
are within noise and are reported as suggestive, not significant. We also flag a selection
caveat: the final configuration ($k{=}10$) was chosen as the best-scoring cell on this same
27-item set, so the headline 0.481 is a max-over-configurations and is optimistically biased
(see \S\ref{sec:disc}).

\begin{table}[t]
\centering
\caption{Final model (SFT+RAG hybrid, $k{=}10$) vs.\ success-criteria gates. Three of four
gates are met; the absolute exact-match target is not.}
\label{tab:results}
\begin{tabular}{lrrl}
\toprule
Metric & Result & Target & Verdict \\
\midrule
Citation exact-match (sec$+$subsec) & 0.481 & $\ge 0.70$ & not met \\
Hallucinated-citation rate          & 0.00  & $\le 0.05$ & met \\
Lift over zero-shot base            & $+0.48$ & $\ge 0.15$ & met \\
Evidenced train-vs-retrieve answer  & yes   & ---        & met \\
\bottomrule
\end{tabular}
\end{table}

\begin{table}[t]
\centering
\caption{Four-arm head-to-head and ablations on the 27-item real eval (single-run point
estimates). SFT-only learns format but mis-recalls sections; RAG is essential and zeroes
hallucination by construction; the hybrid at $k{=}10$ scores highest, though gaps among the
retrieval-bearing rows are within noise at $n{=}27$. The lower block shows that swapping the
cheap \texttt{bge-small} retriever for a bigger, specialized pipeline (a larger embedder and a
cross-encoder reranker) does \emph{not} help either arm (0.37 vs.\ the cheap hybrid's 0.481),
and does not raise recall: the efficiency result. F1 was not computed for the light-SFT and
specialized-retrieval ablations (shown as ``---'').}
\label{tab:arms}
\begin{tabular}{llrrrr}
\toprule
Arm & Config & $k$ & recall@$k$ & Exact & F1 \\
\midrule
base zero-shot    & ---                 & --- & ---  & 0.00  & 0.00 \\
SFT-only          & LoRA r32/400        & --- & ---  & 0.148 & 0.167 \\
\midrule
\multicolumn{6}{l}{\emph{Cheap retrieval (\texttt{bge-small} hybrid BM25$+$dense)}} \\
RAG-only          & $k{=}5$             & 5   & 0.815& 0.44  & 0.52 \\
RAG-only          & $k{=}10$            & 10  & 0.889& 0.37  & 0.46 \\
hybrid            & r32/400, $k{=}5$    & 5   & 0.815& 0.333 & 0.43 \\
hybrid (light)    & r16/150, $k{=}5$    & 5   & 0.815& 0.333 & --- \\
\textbf{hybrid (final)} & \textbf{r32/400, $k{=}10$} & \textbf{10} & \textbf{0.889} & \textbf{0.481} & \textbf{0.574} \\
\midrule
\multicolumn{6}{l}{\emph{Bigger, specialized retrieval (\texttt{bge-large} $+$ cross-encoder reranker)}} \\
RAG-only          & $k{=}10$            & 10  & 0.852& 0.37  & --- \\
hybrid            & r32/400, $k{=}10$   & 10  & 0.852& 0.37  & --- \\
\bottomrule
\end{tabular}
\end{table}

\begin{figure}[t]
\centering
\includegraphics[width=0.72\linewidth]{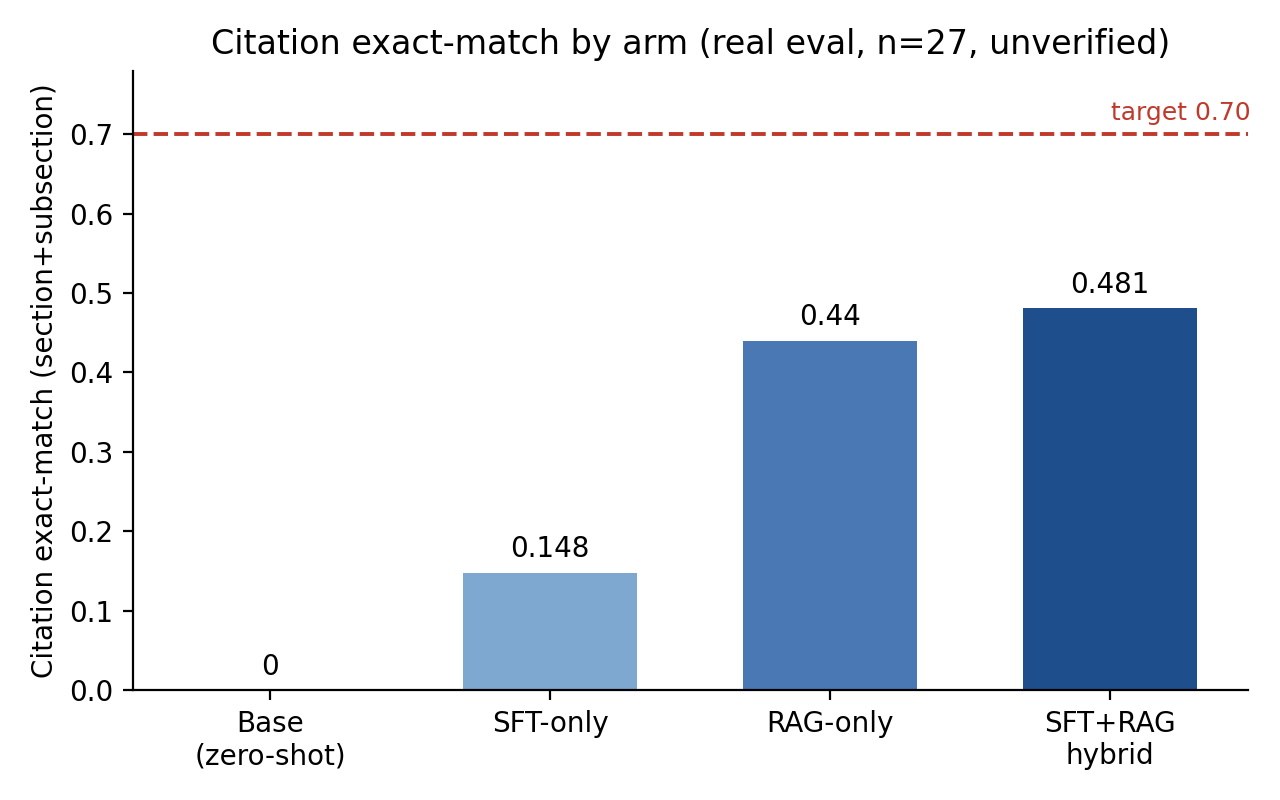}
\caption{Citation exact-match rises monotonically across arms (base 0.00, SFT-only
0.148, RAG-only 0.44, SFT+RAG hybrid 0.481), but the best arm still
falls short of the 0.70 target (dashed). Real eval, $n{=}27$, human-verification-pending.}
\label{fig:arms}
\end{figure}

\paragraph{Train vs.\ retrieve (Fig.~\ref{fig:arms}).} The base model cannot cite the RTA:
0.00 exact-match with 81\% of citations hallucinated. SFT-only teaches the citation
\emph{format} and cuts the hallucination rate from 0.81 to 0.148, but mis-recalls exact
sections, reaching only 0.148 exact-match (the two 0.148 figures are distinct quantities that
coincide), consistent with exact section numbers being a memorization weak spot for
$\sim$8B models \citep{2603.13307v1}. RAG is the decisive lever: RAG-only reaches 0.44 and
drives hallucination to 0.00, which (as noted in \S\ref{sec:results}) follows by
construction from the cite-from-context existence check rather than being learned. The hybrid
scores highest at 0.481 exact-match (0.574 F1, 0.00 hallucination), though its
$\sim$0.04 ($\approx$1-item) margin over RAG-only is within noise at $n{=}27$. The
qualitatively robust finding is the large step from the non-retrieval arms ($\le$0.148) to the
retrieval-bearing arms ($\ge$0.37), not the ordering within the latter.

\paragraph{A candidate-depth ($k$) pattern (hypothesis).} The hybrid's apparent advantage is
conditional on candidate depth. More candidates appear to \emph{hurt} zero-shot RAG (0.44 at
$k{=}5\rightarrow$0.37 at $k{=}10$, even as recall rises 0.815$\rightarrow$0.889), as if
distractors mislead selection; but the same depth \emph{helps} the SFT model (0.333
at $k{=}5\rightarrow$0.481 at $k{=}10$). Our reading is that SFT teaches more
robust selection from a noisier, higher-recall candidate set. We stress two limits on this
claim. First, the deltas are 1--4 items on $n{=}27$, so the difference-in-differences
``interaction'' is under-powered and we offer it as a hypothesis, not an established effect.
Second, it is confounded: recall \emph{also} rises with $k$ (0.815$\rightarrow$0.889), so the
hybrid's $k{=}10$ gain could be additional retrieval supply rather than better selection; the
$\mathrm{recall@}k \times P(\text{select})$ decomposition in Eq.~(1) is motivating, but we did
not measure the conditional selection term $P(\text{select}\mid\text{retrieved})$ directly, so
supply and selection cannot be cleanly separated here. A lighter SFT (r16/150) also scores
0.333 at $k{=}5$, which is consistent with (but, being two single-run points, does not
establish) insensitivity to LoRA strength.

\begin{figure}[t]
\centering
\includegraphics[width=0.72\linewidth]{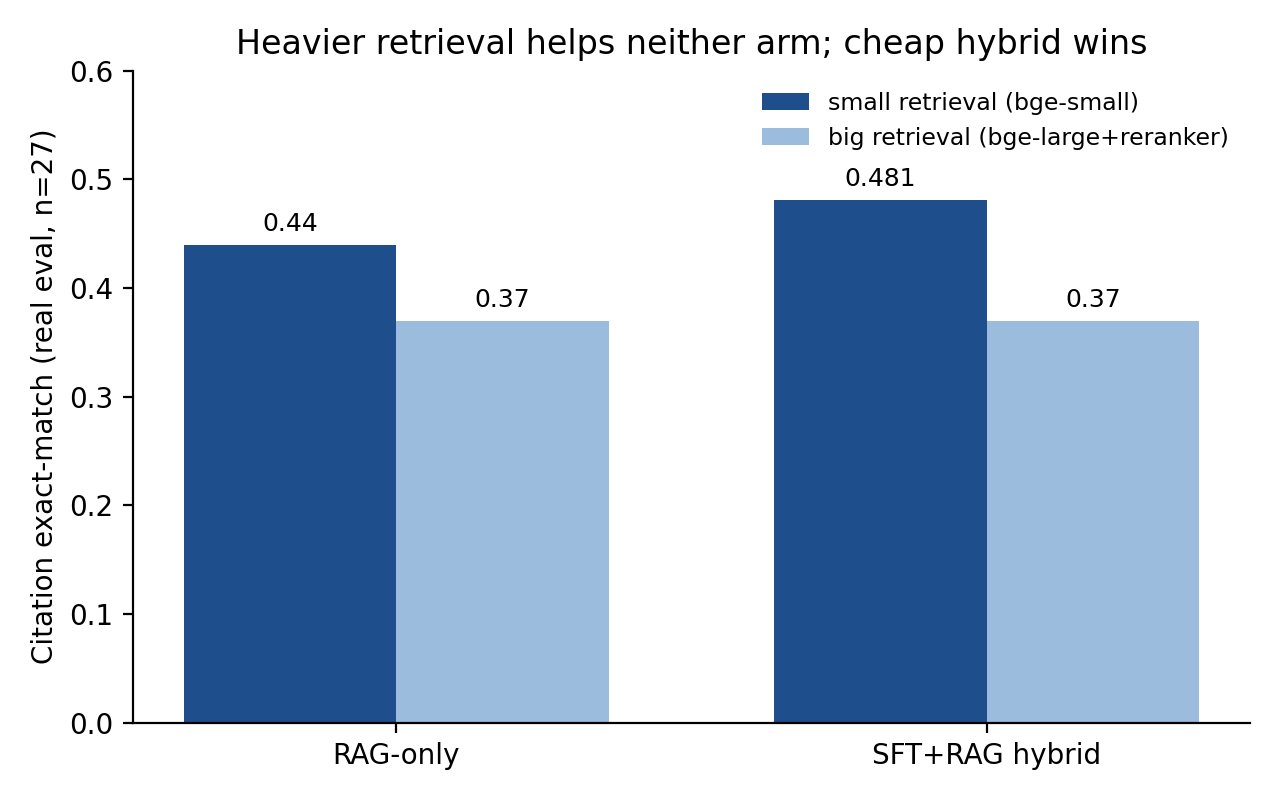}
\caption{Efficiency 2$\times$2: a cheap \texttt{bge-small} hybrid (0.481) is the highest-scoring
cell; a heavier \texttt{bge-large}$+$reranker retriever does not improve either RAG-only (0.37)
or the hybrid (0.37). On a 27-item set these cells carry overlapping uncertainty, so the
reading is that heavier retrieval buys no measurable benefit on this corpus, not that the
small retriever is provably better.}
\label{fig:eff}
\end{figure}

\paragraph{Efficiency: no measurable benefit from heavier retrieval (Fig.~\ref{fig:eff}).} The
cheap \texttt{bge-small} hybrid (0.481) is not beaten by a heavier
retriever (\texttt{bge-large} $+$ cross-encoder reranker), which scores 0.37 for RAG-only
and 0.37 for the hybrid; recall is also no higher (0.889$\rightarrow$0.852).
The observed ordering is hybrid $k{=}10$ (0.481) $\ge$ RAG-only best (0.44) $\ge$ all
big-retrieval cells (0.37). We emphasize that these cells span $\sim$3 items on
$n{=}27$ with overlapping uncertainty: the safe reading is that heavier, specialized retrieval
buys \emph{no measurable improvement} on this short-statute corpus (a useful negative result
for a latency- and cost-constrained deployment), not that the small retriever is provably
superior. Separately, an improved/larger training set (2{,}645 pairs, 63\% subsection-targeted)
did \emph{not} improve over the original, reaching 0.444 with 0.037 hallucination
(again a $\sim$1-item difference); we read the data lever as offering no easy further gain
here, not as exhausted.

\paragraph{Subgroups and errors.} The real slice (0.481) is, surprisingly, \emph{higher} than
the synthetic-test slice (0.327), the opposite of the usual synthetic-overfit worry,
because real questions cluster on a few high-frequency provisions the retriever handles well,
whereas synthetic questions span the long tail. Per-RTA-Part subgroups are not statistically
powered at $n{=}27$ and are deferred. The dominant error mode is retrieval misses
(\textasciitilde11\%; recall@10 $=$ 0.889): when the governing section is not in the
top-10 the model guesses a topically adjacent provision (hallucination stays 0 because the
guess is a \emph{real} section). Secondary modes are mis-selection among retrieved candidates
and subsection drop (section right, subsection missing $\rightarrow$ 0.5 partial credit),
which accounts for much of the gap between F1 (0.574) and exact-match (0.481).

\section{Discussion and Limitations}
\label{sec:disc}
The results give the operator a clear answer at the level the data can support:
\textbf{training alone is insufficient and retrieval is essential.} The hybrid is the
highest-scoring arm and we read its edge as more robust selection from higher-recall candidate
sets, but at $n{=}27$ its margin over RAG-only is within noise, so we do not claim the hybrid
is provably best. The system passes three of four gates: it zeroes hallucinated citations (by
construction) and clears the lift bar by a wide margin (0.481 vs.\ 0.00, target $\ge 0.15$).
A useful efficiency finding is that heavier retrieval and more training data both buy no
measurable improvement on this short-statute corpus, which matters for a latency- and
cost-constrained deployment even though our sample cannot prove the cheap retriever superior.

Honestly, the headline gate is not met: at 0.481 exact-match the system is well below the
aspirational 0.70 target. Three factors bound the gap: a retrieval-recall ceiling around 0.89
(the dominant error mode); subsection-level difficulty (the gap between F1 and exact-match);
and a tiny, human-verification-pending eval set of 27 items. On generalization to untrained
provisions, the only configuration we ran on the unseen-section slice was the improved-data
retrain variant (not the headline model), which scored 0.31 exact-match on a held-out slice of
330 items whose sections were never seen in training; it degrades but does not collapse,
which suggests retrieval can supply untrained provisions, though this evidence is from a
different model than the headline and should be read with that caveat. Calibration is not
applicable to v1: the model emits a citation string with no confidence, so Brier and ECE are
undefined, and it remains future work.

\subsection{Threats to Validity}
\paragraph{Construct and statistical power.} The headline metric rides on a 27-item,
human-verification-pending real eval set. A harness control confirms the scorer and gold are
mechanically correct, but it does not address statistical reliability: with $n{=}27$ and
single-run, no-confidence-interval estimates, one item shifts a rate by $\sim$0.037, so the
fine-grained ordering among the retrieval-bearing arms (the 0.481 vs.\ 0.44 win, the candidate
depth pattern, and the efficiency 2$\times$2) is within noise and is reported as suggestive
only. Two further cautions apply. First, the final configuration was selected on this same
eval set, so the headline 0.481 is a max-over-configurations and is optimistically biased.
Second, the favorable real-vs-synthetic gap (0.481 vs.\ 0.327 on the larger $n{=}150$ synthetic
slice) suggests the real questions cluster on common, high-frequency provisions, so the
headline likely overstates performance on the long tail; the better-powered synthetic number is
arguably the more representative estimate for arbitrary questions. The remedy for all of these,
and the precondition for any claim of statistical certainty, is a larger, topic-balanced,
human-verified eval set, which is our top future-work item. Until then all real-set numbers
should be read as preliminary.
\paragraph{External.} Training is synthetic-heavy (statute-derived paraphrases) and the
study is single-jurisdiction, single-statute (Ontario RTA $+$ one regulation) on a single base
model (Qwen2.5-7B; Llama-3.1-8B not evaluated); legal-NLP components are known not to
port across jurisdictions \citep{2508.00679v1}, so generalization beyond this setting is
unproven.
\paragraph{Internal.} A retrieval-recall ceiling (\textasciitilde0.89) caps achievable
exact-match independent of the generator; leakage is mitigated by group-aware splits
and an unseen-section slice, but the eval set's clustering on high-frequency
provisions is a confound for the headline number.

\section{Conclusion and Future Work}
On the task of citing the correct Ontario RTA provision, a four-arm head-to-head on
Qwen2.5-7B shows that fine-tuning alone is not enough and retrieval is essential. The SFT+RAG
hybrid is the highest-scoring arm (0.481 exact-match, 0.574 F1, zero hallucination by
construction), which we read as SFT making provision selection more robust to the higher-recall
candidate sets that hurt zero-shot RAG, though at $n{=}27$ this margin is within noise rather
than a proven win; on the same data, a heavier retriever and more training data both buy no
measurable improvement. The artifact does not reach the 0.70 exact-match target on a 27-item,
human-verification-pending eval set. Ranked next steps: (1) a larger, topic-balanced,
human-verified real eval set, which is the precondition for treating any fine-grained ranking
as statistically reliable; (2) a stronger retriever or reranker tuned for short statute text,
attacking the dominant retrieval-miss error mode; and (3) subsection-level modeling and
constrained decoding to close the gap between F1 and exact-match, plus citation-confidence
calibration as a stretch goal.\footnote{All artifacts (code, data, the synthetic
question$\rightarrow$citation dataset, and the trained LoRA adapter) are available from the
authors on request. Statute text is from Ontario e-Laws (Government of Ontario, Crown copyright;
reproducible under the King's Printer terms, confirm attribution before redistribution).
Outputs are decision-support, not legal advice.}

\bibliographystyle{unsrtnat}  
\bibliography{references}

\end{document}